\documentclass[runningheads]{llncs}
\usepackage[T1]{fontenc}
\def\doi#1{\href{https://doi.org/\detokenize{#1}}{\url{https://doi.org/\detokenize{#1}}}}
\usepackage{graphicx}
\usepackage{listings}
\usepackage{ upgreek }
\usepackage{amsmath}
\usepackage{amsfonts}
\usepackage{xspace}
\usepackage{float}
\usepackage{wrapfig}
\usepackage{makecell}
\usepackage{multirow}
\usepackage{graphicx}
\usepackage[table,xcdraw]{xcolor}

\newcommand{\ie}{i.e.,~}

\newcommand{\eg}{e.g.,~}

\usepackage{amsmath}

\def\ymean{{\mathbf{\mu^y}}}
\def\ysigma{{\mathbf{\Sigma^y}}}
\def\zmean{{\mathbf{\mu^z}}}
\def\zsigma{{\mathbf{\Sigma^z}}}

\def\sA{{\mathbb{A}}}
\def\sB{{\mathbb{B}}}

\newcommand{\x}{\mathbf{x}}
\newcommand{\y}{\mathbf{y}}
\newcommand{\z}{\mathbf{z}}
\newcommand{\pc}{\mathbf{y}}
\renewcommand{\a}{\mathbf{a}}
\newcommand{\bb}{\mathbf{b}}

\newcommand{\X}{\mathbf{X}}
\newcommand{\Y}{\mathbf{Y}}
\newcommand{\Z}{\mathbf{Z}}

\newcommand{\params}{\Theta}
\newcommand{\mcX}{\mathcal{X}}
\newcommand{\mcY}{\mathcal{Y}}
\newcommand{\mcZ}{\mathcal{Z}}

\newcommand{\mcD}{\mathcal{D}}
\renewcommand{\a}{\mathbf{a}}

\newcommand{\xn}{\x_n}
\newcommand{\yn}{\y_n}
\newcommand{\zn}{\z_n}

\newcommand{\R}[1]{\mathbb{R}^{#1}}

\newcommand{\btheta}{{\boldsymbol{\theta}}}
\newcommand{\bphi}{{\boldsymbol{\phi}}}

\newcommand{\KL}[2]{\operatorname{KL}\left[ #1 \| #2 \right]}
\newcommand{\E}[2]{\mathbb{E}_{#1}\left[#2\right]}

\begin{document}
%

\title{Weakly Supervised Bayesian Shape Modeling from Unsegmented Medical Images}

\titlerunning{Bayesian Image-To-Shape Modeling}
\author{Jadie Adams\inst{1,2}  \and
Krithika Iyer\inst{1,2} \and
Shireen Y. Elhabian\inst{1,2}}
\authorrunning{Adams et al.}
\institute{Scientific Computing and Imaging Institute, University of Utah, UT, USA \and
Kahlert School of Computing, University of Utah, UT, USA \\
\email{ \{jadie, iyerkrithika, shireen\}@sci.utah.edu }
}
\maketitle
\begin{abstract}

Anatomical shape analysis plays a pivotal role in clinical research and hypothesis testing, where the relationship between form and function is paramount. Correspondence-based statistical shape modeling (SSM) facilitates population-level morphometrics but requires a cumbersome, potentially bias-inducing construction pipeline. Traditional construction pipelines require manual and computationally expensive steps, hindering their widespread use. Furthermore, such methods utilize templates or assumptions (\eg linearity) that can bias or limit the expressivity of the variation captured by the constructed SSM. Recent advancements in deep learning have streamlined this process in inference by providing SSM prediction directly from unsegmented medical images. However, the proposed approaches are fully supervised and require utilizing a traditional SSM construction pipeline to create training data, thus inheriting the associated burdens and limitations. 
To address these challenges, we introduce a weakly supervised deep learning approach to predict SSM from images using point cloud supervision. Specifically, we propose reducing the supervision associated with the state-of-the-art fully Bayesian variational information bottleneck DeepSSM (BVIB-DeepSSM) model. BVIB-DeepSSM is an effective, principled framework for predicting probabilistic anatomical shapes from images with quantification of both aleatoric and epistemic uncertainties. Whereas the original BVIB-DeepSSM method requires strong supervision in the form of ground truth correspondence points, the proposed approach utilizes weak supervision via point cloud surface representations, which are more readily obtainable. Furthermore, the proposed approach learns correspondence in a completely data-driven manner without prior assumptions about the expected variability in shape cohort. Our experiments demonstrate that this approach yields similar accuracy and uncertainty estimation to the fully supervised scenario while substantially enhancing the feasibility of model training for SSM construction.
\end{abstract}
\section{Introduction}
Statistical shape modeling (SSM) has emerged as a useful tool in medical imaging and computational anatomy, offering valuable insights into the variability of anatomical structures, such as organs or bones,
across a given population. SSM provides a population-level statistical representation of morphology, enabling wide-ranging applications in clinical research, including disease diagnosis \cite{ambellan2019statistical}, treatment planning \cite{hassan2021automatic}, surgical simulation \cite{haq2020deformable}, and outcome prediction \cite{aldieri2022improving}.
In SSM, shapes are either represented explicitly via landmark or correspondence points, or implicitly via deformation fields (coordinate transformations in relation to a predefined or learnable atlas) \cite{miller2014diffeomorphometry,cootes2004diffeomorphic}.
The point distribution model (PDM) is a widely adopted explicit shape representation consisting of dense sets of correspondence points defined on the surface of the anatomical shapes in semantically consistent locations across the population.
Traditionally, PDMs were automatically defined on preprocessed shape cohorts (segmented from medical images) via pairwise mapping to a predefined or learned atlas/template (e.g., \cite{styner2006spharm}) or via groupwise optimization (e.g., \cite{cates2017shapeworks}).
Such SSM construction pipelines require time-consuming and expert-driven steps such as segmentation, shape registration, and optimization parameter tuning or atlas construction.  Furthermore, each time a new shape is added, the pipeline must be rerun as optimization is performed across the entire cohort simultaneously. 

Deep learning approaches, such as DeepSSM \cite{bhalodia2018deepssm,bhalodia2024deepssm}, offer an alternative to traditional pipelines by leveraging trained neural networks to directly infer PDMs from unsegmented volumetric images with minimal preprocessing. In inference, this alleviated the need for segmentation and reoptimization given a new scan.
However, integrating deep learning-based solutions into clinical practice necessitates understanding the uncertainty associated with model predictions. Therefore, Bayesian deep learning frameworks have been proposed to provide probabilistic PDM predictions capable of quantifying the two primary forms of uncertainty: aleatoric (data-dependent) and epistemic (model-dependent) \cite{adams2020uncertain,adams2022vib,tothova2018uncertainty,adams2023bvib}.
A notable approach is BVIB-DeepSSM \cite{adams2023bvib}, a probabilistic formulation of DeepSSM \cite{bhalodia2024deepssm} that utilizes a fully Bayesian extension of the variational information bottleneck (VIB) framework \cite{alemi2016deepvib}. 
BVIB-DeepSSM provides PDM prediction from unsegmented images with estimates of aleatoric and epistemic uncertainty that correlate with prediction error, ensuring reliable prediction without compromising accuracy.

Even though deep learning approaches mitigate the overhead associated with SSM construction during inference, they still depend on traditional SSM techniques to construct image/PDM pairs to supervise network training. 
This reliance not only slows down the training preparation process but also means that the network inherits any limiting assumptions made during the construction of training PDMs. Such biases or assumptions can arise from various sources, such as atlas selection in pairwise surface matching approaches or in the definition of optimization objectives.
For instance, the current state-of-the-art (SOTA) groupwise optimization PDM construction method, known as particle-based shape modeling (PSM) \cite{cates2017shapeworks,goparaju2022benchmarking}, imposes a linearity assumption. This assumption restricts the ability of PSM to accurately represent complex, nonlinear shape variations. Training networks on PSM-constructed PDMs could similarly bias network predictions. 

We propose leveraging weak supervision from point clouds in BVIB-DeepSSM training to overcome these limitations. Point cloud shape representations consist of sets of unordered, nonuniform points that sample the surface of the shape. Recently, there has been growing interest in learning SSM from point clouds due to their ease of acquisition compared to the complete, noise-free surface representations (such as meshes or binary volumes) required by traditional SSM construction methods \cite{adams2023pcn,adams2023point2ssm}.
This work proposes training BVIB-DeepSSM using image/point cloud pairs instead of image/PDM pairs. This approach significantly reduces the required supervision and enables training the model on readily available segmentation datasets. 
Our contributions are summarized as follows:
\begin{itemize}
    \item We provide a framework to predict SSM directly from images with reduced supervision by utilizing point cloud shape representations in training rather than ground truth PDMs. 
    \item We introduce formulations of the VIB and fully Bayesian VIB objectives that utilize permutation-invariant Chamfer distance. 
    \item We provide comprehensive experiments that demonstrate that the proposed approach improves the feasibility of predicting SSM from images without sacrificing accuracy or uncertainty calibration. 
\end{itemize}
\section{Related Work}
Traditional PDM construction methods utilize metrics such as entropy \cite{cates2007shape} or minimum description length \cite{davies2002minimum}, or employ parametric representations \cite{ovsjanikov2012functional,styner2006framework,nain2007statistical}.
PSM \cite{cates2017shapeworks} represents the SOTA optimization-based technique for group-wise SSM construction \cite{goparaju2022benchmarking}. However, PSM assumes linear correlations, leading to a bias in the captured population variation. 

DeepSSM \cite{bhalodia2018deepssm,bhalodia2024deepssm} was the pioneering deep learning approach to predict PDMs directly from raw, unsegmented images. DeepSSM utilizes PDM supervision, where training labels are constructed via the full PSM pipeline (including segmentation, preprocessing and alignment, and PDM optimization). 
Uncertain-DeepSSM \cite{adams2020uncertain} adapted the DeepSSM network to be Bayesian, providing aleatoric and epistemic uncertainties.
DeepSSM, Uncertain-DeepSSM, and other formulations \cite{tothova2018uncertainty} rely on a supervised low-dimensional encoding (\ie shape descriptors), precomputed using principal component analysis (PCA). PCA supervision enforces a linear relationship between the latent and the output spaces and restricts the learning task to strictly SSM prediction. Additionally, PCA does not scale in the case of large sets of high-dimensional shape data.
In contrast, VIB-DeepSSM \cite{adams2022vib} introduced a variational information bottleneck (VIB) architecture \cite{alemi2016deepvib} to learn a low-dimensional latent encoding tailored to the PDM estimation task, leading to improved generalization and more accurate estimation of aleatoric uncertainty. 
However, the VIB framework is only half Bayesian \cite{alemi2020half}; thus, VIB-DeepSSM lacks the capability to quantify epistemic uncertainty.
BVIB-DeepSSM \cite{adams2023bvib} extended the VIB-DeepSSM framework to be fully Bayesian, enabling the prediction of probabilistic shapes from images with quantification for both forms of uncertainty. This SOTA model is the basis of the proposed approach.

Recent work has explored unsupervised estimation of SSM from various shape representations \cite{adams2023pcn,adams2023point2ssm,iyer2023mesh2ssm,iyer2024scorp}. 
One study demonstrated that networks designed for point cloud competition perform reasonably well at the task of anatomical PDM generation \cite{adams2023pcn}. These networks typically have an encoder-decoder architecture with a bottleneck \cite{fei2022pcnsurvey}.
The decoder provides a continuous mapping from the learned latent space to output space, resulting in consistently ordered output point clouds, providing correspondence as a by-product.
Mesh2SSM \cite{iyer2023mesh2ssm} explicitly predicts PDMs in an unsupervised manner from mesh shape representations, utilizing complete surface information. 
Point2SSM \cite{adams2023point2ssm} is a self-supervised technique proposed to predict anatomical SSM from point clouds. By employing Chamfer distance reconstruction loss, Point2SSM encourages the predicted PDMs to accurately sample the entire point clouds. 
Recently, SCorP \cite{iyer2024scorp} proposed leveraging a shape prior learned from surface meshes to predict PDMs from unsegmented images within a student/teacher framework. 
While closely related to the proposed method, SCorP lacks uncertainty quantification and necessitates complete mesh surface representations. 
The proposed method requires only image/point cloud pairs to supervise network training, providing PDM prediction and granular uncertainty estimates in inference.
\section{Background}
\subsection{Notation}
Let $\X$, $\Y$, and $\Z$ denote random variables and let $\x$, $\y$, and $\z$ denote realizations of those respective random variables. 
Given an unsegmented volumetric image of an anatomy, denoted  $\x \in \mathbb{R}^{H \times W \times D}$, the goal is to predict a PDM denoted $\hat{\y} \in \mathbb{R}^{3M}$. Each PDM is a set $M$ \textit{ordered correspondence points}, where a 3D vector of coordinates defines each correspondence point. 
Training the network requires a set of paired data, denoted $\mathcal{D} = \{ \mcX, \mcY \}$. 
Here $\mcX = \{\xn\}_{n=1}^N$ is a set of $N$ unsegmented volumetric images.
In previously proposed fully supervised settings, $\mcY = \{\yn\}_{n=1}^N$ where $\yn \in \mathbb{R}^{3M}$ denotes a ground truth PDM, constructed via a traditional pipeline, comprised of $M$ \textit{ordered correspondence points}.
In the proposed weakly supervised setting, $\pc_n$ denotes a point cloud shape representation, meaning an \textit{unordered set} of points on the surface of the shape $n$. 
The VIB framework utilizes a learned stochastic latent encoding: $\mcZ = \{\zn \}_{n=1}^{N}$, where $\zn \in \R{L}$ and $L \ll 3M$.

\subsection{Variational Information Bottleneck}
In information bottleneck (IB) theory, a stochastic encoding $\Z$ is learned to capture the minimal sufficient statistics required of input $\X$ to predict the output $\Y$ \cite{tishby2000information}.  
The encoding $\Z$ and model parameters $\params$ are estimated by maximizing the IB objective: 
    \begin{equation}\label{eq:IB}
        \operatorname{argmax}_\params I(\Y,\Z; \params) - \beta I(\X, \Z; \params) 
    \end{equation}
where $I$ denotes mutual information and $\beta$ is a Lagrangian multiplier. 

The first term in Eq. \ref{eq:IB} encourages $\Z$ to be maximally expressive of $\Y$, encouraging predictive accuracy. The second term encourages $\Z$ to be maximally compressive of $\X$, affecting the model complexity. 

In the deep variational information bottleneck (VIB) \cite{alemi2016deepvib} approach, the IB model is parameterized via a neural network with weights $\params = \{\bphi, \btheta\}$, and a latent distribution is learned by minimizing the IB objective (Eq. \ref{eq:IB}). Direct calculation of mutual information is intractable in this context, and thus VIB employs variational inference to derive a theoretical lower bound on the IB objective:
\begin{equation}
    \mathcal{L}_{VIB} = \E{\hat{\Z} \sim q(\Z|\X, \bphi)}{- \log p(\Y|\hat{\Z}, \btheta)} + \beta\KL{q(\Z|\X, \bphi)}{p(\Z)}
    \label{eq:VIB_elbo}
\end{equation}
The first term, the negative log-likelihood (NLL), encourages $\Z$ to be predictive of $\Y$. The second term, the Kullback–Leibler (KL) divergence, encourages $\Z$ to be compressive of $\X$. The $\beta$ hyper-parameter controls the tradeoff.

\subsection{BVIB-DeepSSM}
\begin{figure}[!h]
    \begin{center}
    \includegraphics[width=.9\textwidth]{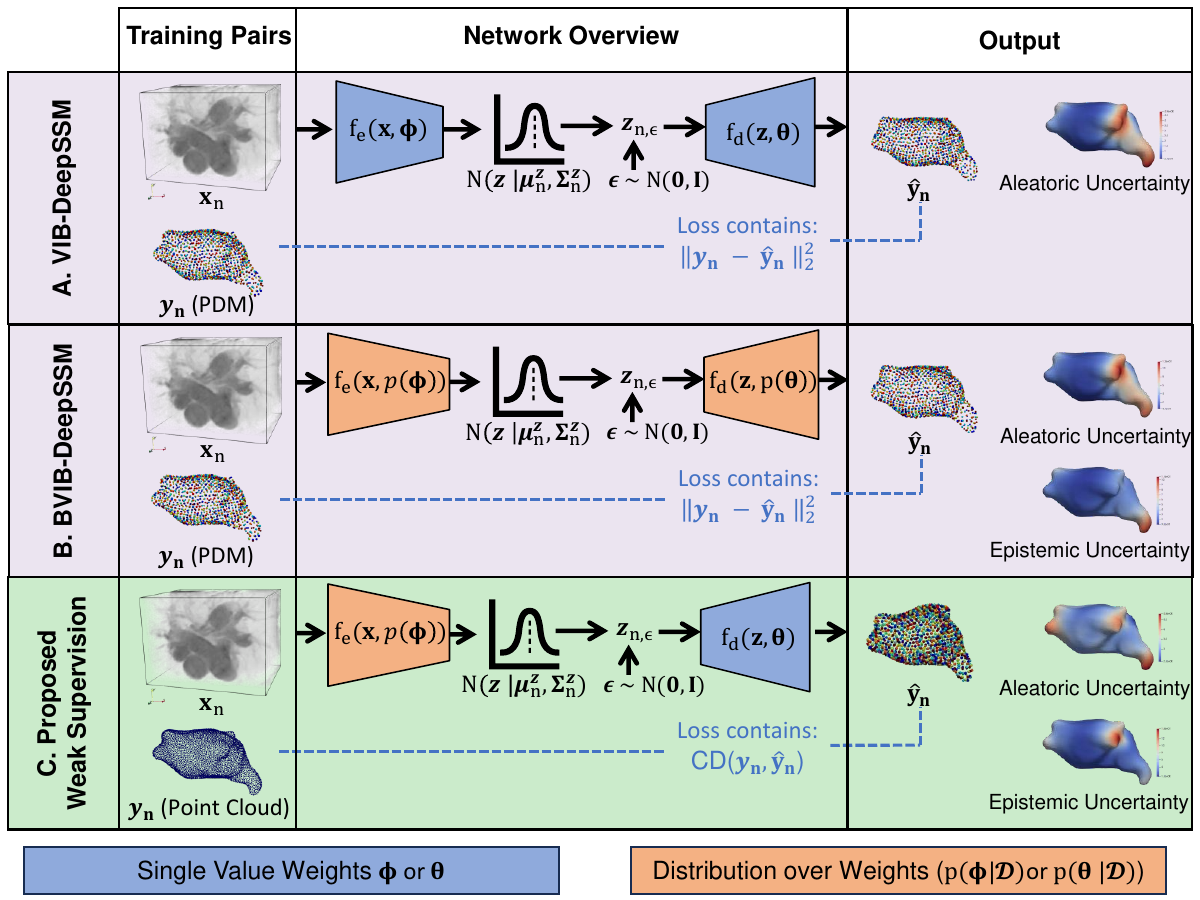}
    \end{center}
    \caption{Overview of the differences between VIB-DeepSSM \cite{adams2022vib}, BVIB-DeepSSM \cite{adams2023bvib}, and the proposed weakly supervised variant of BVIB-DeepSSM with point cloud supervision.}
    \label{fig:arch}
\end{figure}

VIB-DeepSSM \cite{adams2022vib} employs the VIB \cite{alemi2016deepvib} approach to learn the latent encoding in the context of the task: predicting PDM $\hat{\y}$ from image $\x$. 
The VIB-DeepSSM architecture (Fig. \ref{fig:arch}.A) is comprised of an encoder and decoder.  The encoder, $f_e$, comprised of 3D convolutional and densely connected layers parameterized by $\bphi$, maps the input image to a Gaussian latent distribution: $\mathcal{N}(\z|\zmean,\zsigma)$.
Posterior samples $\z_{\epsilon}$ are acquired from this predicted latent distribution using the reparameterization trick to enable gradient calculation.
The decoder, $f_d$ parameterized by $\btheta$, maps the latent encoding to the predicted output $\hat{\y}$.

VIB-DeepSSM allows for capturing aleatoric uncertainty as the variance of the $p(\y|\z)$ distribution.
This variance is computed by sampling multiple latent encodings from $\mathcal{N}(\z|\zmean,\zsigma)$ and passing them through the decoder to get a sampled distribution of predictions. A Gaussian distribution is estimated from these samples denoted: $\mathcal{N}(\hat{\y}|\ymean,\ysigma)$. The estimated $\ysigma$ captures the aleatoric or data-dependent uncertainty.
However, this approach does not quantify epistemic uncertainty because VIB is only half-Bayesian \cite{alemi2020half}.

BVIB-DeepSSM \cite{adams2023bvib} derived the fully Bayesian VIB formulation by applying an additional PAC-Bound with respect to the network parameters. 
In the VIB-DeepSSM model, parameters $\params = \{\bphi, \btheta\}$ are fit via maximum likelihood estimation. BVIB-DeepSSM utilizes variational inference to approximate the posterior $p(\params|\mcD)$. The BVIB-DeepSSM objective results in two intractable posteriors $p(\Z|\X,\bphi)$ and $p(\params|\X, \Y)$, the former is approximated via $q(\Z|\X, \bphi)$ as in Eq. \ref{eq:VIB_elbo} and the latter is approximated by $q(\params)$. The two KL divergence terms are minimized via a joint evidence lower bound, resulting in the objective:

\begin{align}
    \mathcal{L}_\text{BVIB} &= \E{\tilde{\params}}{\E{\hat{\Z}}{- \log p(\Y|\hat{\Z}, \tilde{\btheta})} +
     \beta \KL{q(\Z|\X, \tilde{\bphi})}{p(\Z)}} 
     + \KL{q(\params)}{p(\params)}
    \label{eq:BVIB_loss}
\end{align}

In BVIB-DeepSSM (Fig. \ref{fig:arch}.B), epistemic uncertainty is captured as the variance in predictions made with various network weights sampled from the learned distribution $p(\params|\mcD)$. 
Many approaches have been proposed for the computationally challenging task of estimating a distribution over model parameters for epistemic uncertainty quantification. Among these approaches, concrete dropout and ensembling have been shown to be most effective \cite{adams2023benchmarking} and are utilized in the BVIB-DeepSSM formulation \cite{adams2023bvib}. 
Concrete dropout employs Monte Carlo dropout sampling as a practical approach for approximate variational inference \cite{gal2017concrete}. This approach utilizes a continuous relaxation of the Bernoulli distribution (\ie concrete distribution) to parameterize the learned distribution over weights. Epistemic uncertainty is captured by the spread of predictions resulting from inference with various sampled dropout masks. Concrete dropout automatically optimizes the dropout probabilities at each layer alongside the network weights.
Batch ensemble \cite{wen2020batchensemble} compromises between a single network and a full deep ensemble, achieving a balance between performance, computation time, and memory usage. 
BVIB-DeepSSM additionally proposed combining concrete dropout and batch ensemble to acquire a multimodal approximate posterior on weights for increased flexibility and expressiveness \cite{adams2023bvib}.

\section{Methods}
We propose using point cloud shape representations to weakly supervise BVIB-DeepSSM.
Recent work has shown that bottleneck network architectures with fixed decoders supervised by point cloud-based loss can learn correspondence \cite{adams2023pcn}. In such networks, the bottleneck captures a population-specific shape prior. Directly decoding the latent shape feature representation results in a consistent ordering of the output point clouds across samples, providing PDMs.
We propose to leverage this effect in BVIB-DeepSSM, allowing for the replacement of ground truth PDMs with unordered point clouds in the training data. This advancement requires updating the BVIB-DeepSSM formulation in two crucial ways: first, the objective must altered for point cloud supervision, and second, consideration regarding the epistemic uncertainty quantification approach must be made. 

\subsection{Proposed Weakly Supervised Loss}
Reducing the supervision requires updating the first term in the VIB and BVIB-DeepSSM objectives (Eq. \ref{eq:BVIB_loss} and \ref{eq:BVIB_loss}), the NLL term. 
In the PDM-supervised setting, the NLL term is expressed as:
\begin{equation}
   - \log p(\y|\ymean, \ysigma) = \frac{|| \ymean - \y||_2}{2\ysigma} + \frac{1}{2}\log{\ysigma}
   \label{eq:NLL}
\end{equation}
where $\ymean$ and $\ysigma$ are the mean and variance of the predicted distribution estimated using various posterior samples $\tilde{\params}$ and $\hat{\z}$. 

Replacing $\y$ (an ordered PDM) with unordered point clouds requires replacing the L2 norm in Eq. \ref{eq:NLL} with a permutation invariant distance metric.
Chamfer distance is most commonly used for this purpose. The Chamfer distance from point cloud $\sA$ to point cloud $\sB$ is defined as: 
\begin{equation}
    \mathrm{CD}(\sA \rightarrow \sB) = 
        \frac{1}{|\sA|} \sum_{\a \in \sA} \min_{\bb \in \sB} || \a - \bb ||_2^2 
        \label{eq:CD}
\end{equation} 
Typically, bidirectional Chamfer distance is used:
\begin{equation}
    \mathrm{CD}(\sA, \sB) =  \mathrm{CD}(\sA \rightarrow \sB) + \mathrm{CD}(\sB \rightarrow \sA)
    \label{eq:full_CD}
\end{equation}
Note the number of points in $\sA$, denoted $|\sA|$, is not required to match $|\sB|$.

We propose utilizing the single directional distance $\mathrm{CD}(\ymean \rightarrow \pc)$ as a replacement for the L2 norm in Eq. \ref{eq:NLL}, as this is a commensurate metric that is calculated point-wise, in a permutation-invariant manner. 
The resulting updated NLL term is expressed as:
\begin{equation}
   - \log p(\pc |\ymean, \ysigma) \approx \frac{\mathrm{CD}(\ymean \rightarrow \pc )}{2\ysigma} + \frac{1}{2}\log{\ysigma}
   \label{eq:NLL_PC}
\end{equation}
This update enables permutation invariant point-wise error estimation. However, the single-directional $\mathrm{CD}$ does not ensure that the predicted PDM will sample the entire surface well. For instance, if all points in $\ymean$ converge to a single point in $\pc$, $\mathrm{CD}(\ymean \rightarrow \pc)$ would be minimized. To prevent this behavior, we include $\mathrm{CD}(\pc \rightarrow \ymean)$ as a regularization term. This term encourages the predicted points to be well-spread across the surface so that each point in $\pc$ has a close neighbor in $\ymean$. The resulting updated VIB-DeepSSM objective is expressed as:
Thus, the proposed BVIB-DeepSSM objective is expressed as:
\begin{equation}
    \mathcal{L}_\text{Proposed VIB} = {\E{\hat{\Z}}{- \log p(\Y|\hat{\Z}, \btheta)} + \beta
     \KL{q(\Z|\X, \bphi)}{p(\Z)}}  + \alpha \mathrm{CD}(\Y \rightarrow \hat{\Y})
    \label{eq:CD_VIB_loss}
\end{equation}
where $\log p(\Y^|\hat{\Z}, \btheta)$ is computed via Eq. \ref{eq:NLL_PC} and $\mathrm{CD}(\Y \rightarrow \hat{\Y})$ is the regularization term computed via Eq. \ref{eq:CD}, weighted by hyperparameter $\alpha$.

\subsection{Weakly Supervised Epistemic Uncertainty Quantification}

In addition to updating the learning objective, the reduction in supervision requires adapting the approach to epistemic uncertainty quantification. While ensembling proved to be an effective frequentist approximation for estimating a distribution over model parameters in the original BVIB-DeepSSM formulation, it is not appropriate in the weakly supervised setting. This is because point cloud supervision does not enforce one particular point ordering in correspondence prediction as PDM supervision does. Rather, network-specific correspondence is induced by two factors: the Chamfer distance reconstruction loss and the consistent, continuous mapping from the latent space to the output space provided by the decoder. Thus, while a given network provides correspondence across predictions, there is no mechanism to enforce correspondence consistency across different networks or ensemble members. Each member would learn a unique output point ordering, rendering the ensemble averaging effect meaningless. Thus, in the weakly supervised context, a true Bayesian approximation method must be used to learn a distribution over weights within a single network. 

Additionally, introducing stochasticity to the decoder would be detrimental to PDM prediction, as correspondence is induced by the established continuous mapping from the latent to output space. Thus, we propose adapting the concrete dropout-based BVIB-DeepSSM model to estimate epistemic uncertainty from the encoder alone and utilize a fully deterministic decoder. Here, predictive distributions are acquired by decoding various $\Z$ samples with various encoder dropout masks. 

Thus, the proposed BVIB-DeepSSM objective is expressed as:
\begin{align}
    \mathcal{L}_\text{Proposed} &= \E{\tilde{\bphi}}{\E{\hat{\Z}}{- \log p(\Y|\hat{\Z}, \btheta)} + \beta
     \KL{q(\Z|\X, \tilde{\bphi})}{p(\Z)}} \nonumber \\
     &+ \KL{q(\bphi)}{p(\bphi)} +  \alpha \mathrm{CD}(\Y \rightarrow \hat{\Y})
    \label{eq:CD_BVIB_loss}
\end{align}
where, as in Eq. \ref{eq:CD_VIB_loss}, $\log p(\Y|\hat{\Z}, \btheta)$ is computed via Eq. \ref{eq:NLL_PC}.

An overview of VIB-DeepSSM, BVIB-DeepSSM, and the proposed weakly supervised BVIB-DeepSSM approach are provided in Fig. \ref{fig:arch}.

\section{Experiments}

\subsection{Datasets}
We utilize two challenging datasets to evaluate the proposed method: the left atrium and the liver.
The left atrium dataset includes 1,096 shapes derived from cardiac late gadolinium enhancement MRI images of different atrial fibrillation patients. The images were manually segmented at the University of Utah Division of Cardiovascular Medicine with spatial resolution $0.65 \times 0.65 \times 2.5$ mm$^3$, and the endocardium wall was used to cut off pulmonary veins. This dataset includes substantial morphological diversity in overall size, the size of the left atrium appendage, and the quantity and arrangement of pulmonary veins. Following BVIB-DeepSSM, we hold out outlier cases in the test set, selected via thresholding on an outlier degree computed on images and meshes \cite{moghaddam1997probabilistic}. The resulting test set includes 40 shape outliers, 78 image outliers, and 92 randomly selected inlier test samples.
The liver dataset contains 834 CT scans and corresponding quality-controlled segmentations from the open-source AbdomenCT-1K dataset  \cite{Ma-2021-AbdomenCT-1K}. These images vary significantly in intensity, quality, and resolution, providing a challenging test case. We randomly split the liver data 80$\%$/10$\%$/10$\%$. into training, validation, and test sets.

\subsection{Experimental Setup}
We compare the proposed weakly supervised adaptations of VIB-DeepSSM and BVIB-DeepSSM with the original formulations. 
We utilize the PSM construction method implemented in the ShapeWorks software suite \cite{cates2017shapeworks} to create PDMs for training fully supervised methods. Additionally, we utilize ShapeWorks to process images and segmentations, including cropping around the region of interest and downsampling to manage memory usage.
We generate surface meshes with 5000 vertices from the segmentations. The vertices serve as point clouds for the proposed weak supervision.
We employ image augmentation in training all models in the form of additive Gaussian noise with random variance between 0 and 1\% of the full signal.
Following the BVIB-DeepSSM strategy, burn-in is used to convert the loss from deterministic (L2 or CD) to probabilistic (Eqs \ref{eq:BVIB_loss} and \ref{eq:CD_BVIB_loss}) \cite{adams2023bvib}. 
This burn-in counteracts the accuracy reduction that occurs when NLL-based loss is used with a gradient-based optimizer \cite{seitzer2021pitfalls}.
The concrete dropout implementation of BVIB-DeepSSM is used with initial dropout probabilities of 0.1.
All models were trained until the validation error (either L2 or CD, depending on supervision) had not decreased in 50 epochs.
The training was done on Tesla V100 GPU with Xavier initialization \cite{xavier2010initialization}, Adam optimization \cite{adam}.
Full model parameters and training and evaluation code are provided at \url{https://github.com/jadie1/Weakly-Supervised-BVIB-DeepSSM/}.

\subsection{Evaluation Metrics}

There are three factors to consider when evaluating probabilistic PDM prediction accuracy. 
The first is surface sampling accuracy, which assesses how well the points are constrained to capture the complete shape surfaces.
The second is the assessment of how well the population-level statistics are captured through predicted correspondences.
The third is the calibration of the uncertainty estimates. This section describes the metrics used to assess these three factors.

\subsubsection{Surface Sampling:}
A small \textbf{Chamfer distance} between the point cloud and predicted PDM, CD($\hat{\y}, \pc$) (Eq. \eqref{eq:full_CD}), indicates the output points accurately capture the complete shape. \textbf{Point-to-surface distance (P2S)} assesses how well points are constrained to the surface. P2S quantifies the distance of the predicted points to a complete ground truth surface shape representation (\ie mesh).

\subsubsection{Correspondence/SSM Metrics:}
Principal component analysis (PCA) is used in SSM analysis to understand the modes of variation in the population and to evaluate how effectively population-level statistics are captured \cite{munsell2008evaluating}. Three key metrics help measure this: compactness, generalization, and specificity.
\textbf{Compactness (Comp.)}: Good correspondence leads to a more compact SSM, meaning the training data distribution can be represented using a minimal number of parameters. Strong correspondence allows a larger proportion of explained population variance to be captured with fewer PCA modes. A larger area under the cumulative variance plot indicates better correspondence.
\textbf{Generalization (Gen. CD)}: A precise SSM should generalize effectively from training subjects to new, unseen subjects. The generalization metric measures the Chamfer distance (CD) between estimated correspondences from test point clouds and their reconstructions from training SSM-based PCA embeddings using varying numbers of components. A smaller CD indicates better generalization.
\textbf{Specificity (Spec. CD)}:  Specificity assesses whether the predicted SSM produces valid instances of the shape class. It is calculated as the average Chamfer distance between training examples and generated samples from the training SSM-based PCA embeddings using different numbers of components. A smaller CD suggests the SSM is more specific.
Recent work also utilizes mapping error \textbf{(ME)} to estimate correspondence accuracy \cite{adams2023point2ssm}.  
ME quantifies how consistent output point neighborhoods are across the population \cite{lang2021dpc}. A lower ME indicates consistent neighborhoods, implying better correspondence.

\subsubsection{Uncertainty Calibration}
We expect well-calibrated uncertainty estimates to correlate with the error. Thus, a higher Pearson \textbf{r} coefficient between predicted uncertainty and P2F error suggests better calibration. Furthermore, accurate uncertainty estimation is useful in out-of-distribution (OOD) detection.

\subsection{Results}

\begin{figure*}[!ht]
    \begin{center}
    \includegraphics[width=.85\textwidth]{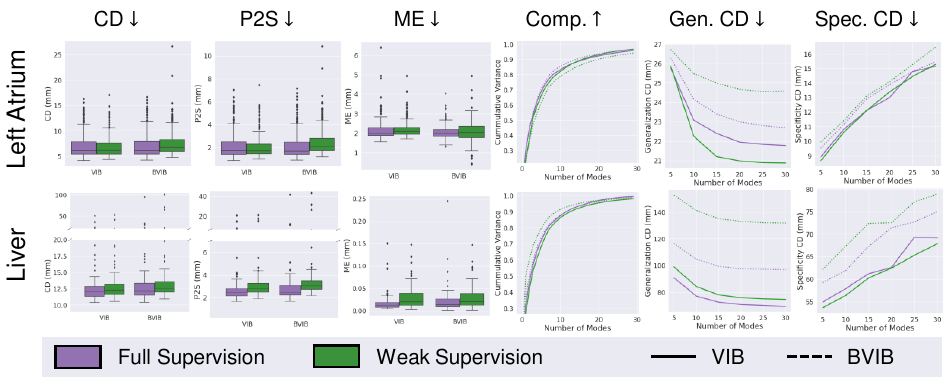}
    \end{center}
    \caption{Results of VIB-DeepSSM and C-BVIB-DeepSSM with PDM-supervision and the proposed point cloud supervision on the left atrium and liver dataset. Box plots show the distribution of errors across the test set. The SSM metric plots show the values with various numbers of PCA modes, where the line type depicts the type of DeepSSM model. Lower values are better for all metrics with the exception of compactness. }
    \label{fig:results}
\end{figure*}

Fig. \ref{fig:results} provides the surface sampling and correspondence evaluation metrics across both test sets. \textbf{The proposed weakly supervised models provide similar accuracy across all metrics while significantly reducing the supervision requirement.} Hence, we are not sacrificing accuracy, but we democratize building networks that provide PDMs directly from unsegmented images by requiring only the level of supervision needed to train segmentation networks.

\vspace{-.4in}

\begin{figure*}[!ht]
    \begin{center}
    \includegraphics[width=.85\textwidth]{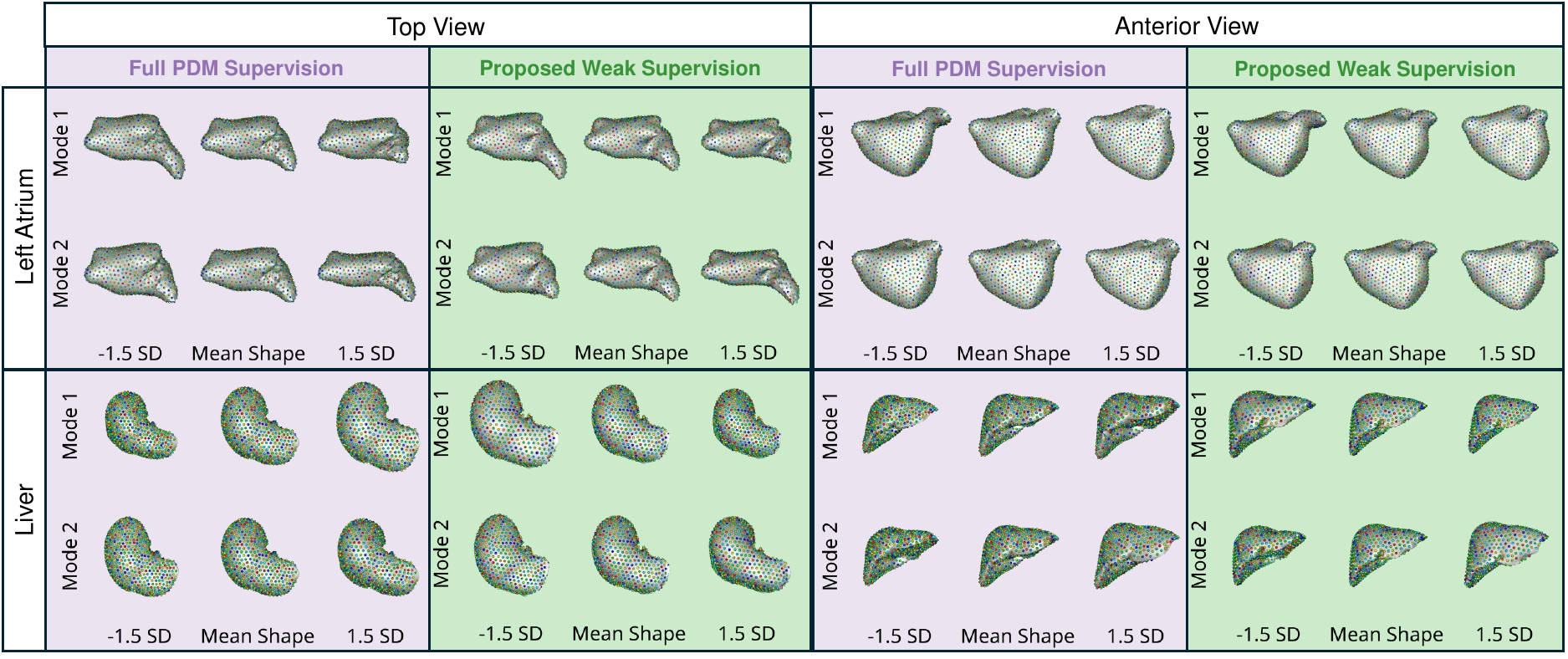}
    \end{center}
    \caption{Modes of variation resulting from the predicted SSM on the test set with the BVIB-DeepSSM models from the top and anterior view. The mean shape is shown with the primary and secondary PCA modes of variation at $\pm 1.5$ standard deviations (SD). Correspondence points are displayed over meshes constructed from the points.}
    \label{fig:modes}
\end{figure*}

Fig. \ref{fig:modes} displays the modes of variation resulting from the predicted SSM on the test sets for the BVIB-DeepSSM model. The mean shape and primary and secondary modes of variation resulting from PDM supervision and the proposed PC supervision are very similar. In the left atrium case, the primary mode captures the length of the left atrium appendage, and the secondary mode captures the volume or sphericity. The primary and secondary modes of variation in the liver dataset capture the size and curvature. 
The results demonstrate that the proposed weak supervision does not lead to less accurate PDM prediction. The predicted points sample the true surface to the same degree and offer a similar correspondence accuracy despite the absence of ground truth correspondence supervision. Additionally, the similarity in the captured modes of variation suggests that PDM predictions made with weak supervision could be equally useful in downstream clinical tasks as the predictions made with full PDM supervision.

 \begin{figure*}[!h]
    \begin{center}
    \includegraphics[width=.8\textwidth]{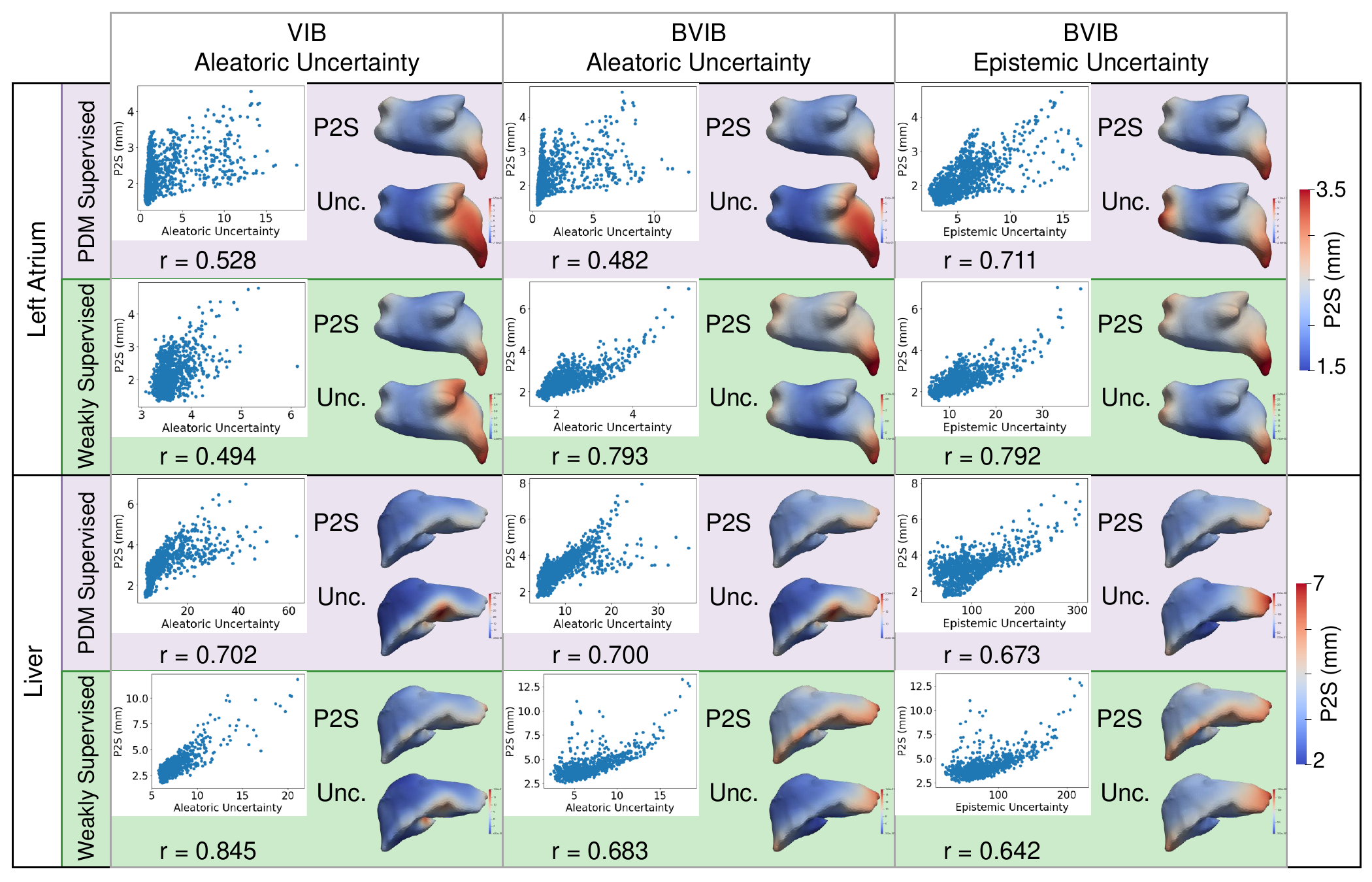}
    \end{center}
    \caption{Uncertainty calibration results. Scatter plots and corresponding Pearson R correlation coefficients demonstrate the point-wise correlation between the estimated uncertainty and P2S error across the test sets. The average P2S error and uncertainty values are also shown via heatmaps on a representative mesh, illustrating spatial correlation.}
    \label{fig:uncertainty}
\end{figure*}
\vspace{-.1in}

Fig. \ref{fig:uncertainty} displays the point-wise correlation between the predicted uncertainty values and P2S distance error across the test set. We would expect the uncertainty estimates to be higher for predicted correspondence points that are further from the true shape surface. The r correlation coefficients and resulting average uncertainty heatmaps are very similar, indicating that reducing the supervision does not significantly impact the uncertainty calibration. These results demonstrate the effectiveness of estimating the NLL term with CD rather than L2 Euclidean distance (Eq. \ref{eq:NLL_PC}). The spatial correlation between the P2S error and uncertainty heatmaps demonstrates the utility of these probabilistic frameworks in aiding in assessing prediction reliability.

\begin{figure*}[!h]
    \begin{center}
    \includegraphics[width=.8\textwidth]{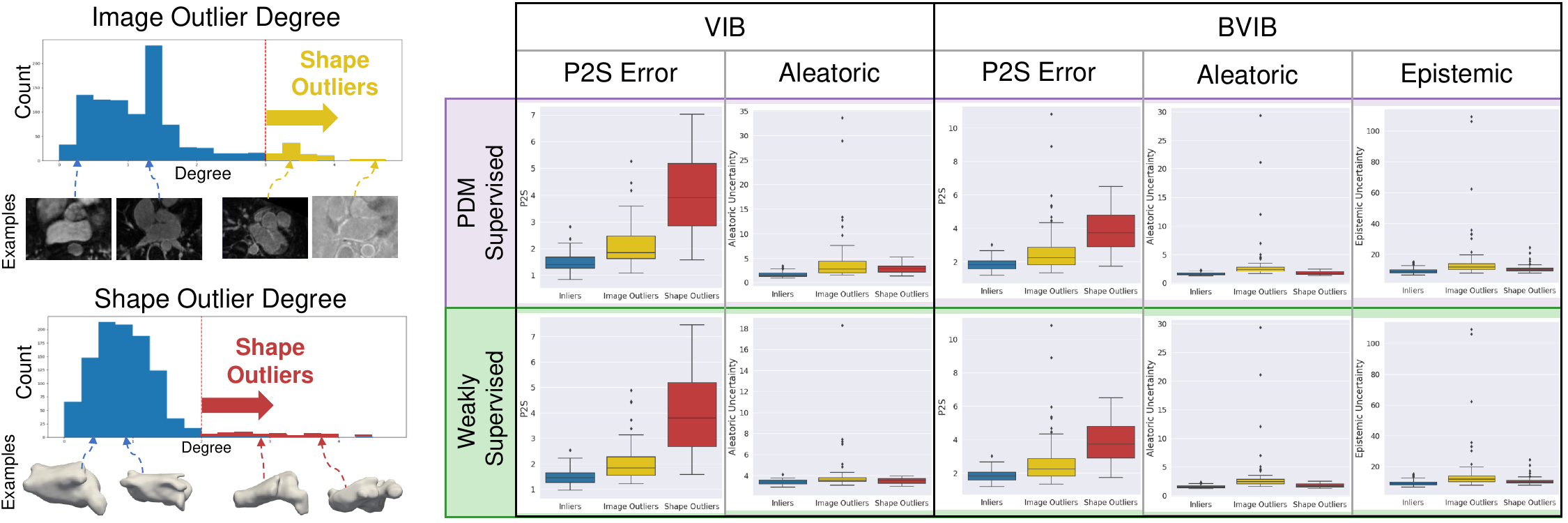}
    \end{center}
    \caption{Left atrium outlier test sets. The histogram plots the distribution of image and shape outlier degrees with example image slices and meshes. Box plots show the distribution of P2S error and uncertainty across the three test sets: inliers (blue), image outliers (yellow), and shape outliers (red).}
    \label{fig:LA_outliers}
\end{figure*}

Uncertainty quantification is also useful in detecting out-of-distribution (OOD) samples. The left atrium dataset is comprised of three subsets: image outliers, shape outliers, and randomly selected inlier examples. This partitioning was performed by thresholding on a precomputed outlier degree \cite{moghaddam1997probabilistic} as shown in Fig. \ref{fig:LA_outliers}. The error and uncertainty estimation distributions across subsets are shown in  Fig. \ref{fig:LA_outliers}. The predicted uncertainty is slightly higher for the outlier test sets, especially for the extreme image outliers. The full and weakly supervised models provide similar patterns in error and uncertainty across the left atrium test sets.

\vspace{-.4in}

 \begin{figure*}[!h]
    \begin{center}
    \includegraphics[width=.8\textwidth]{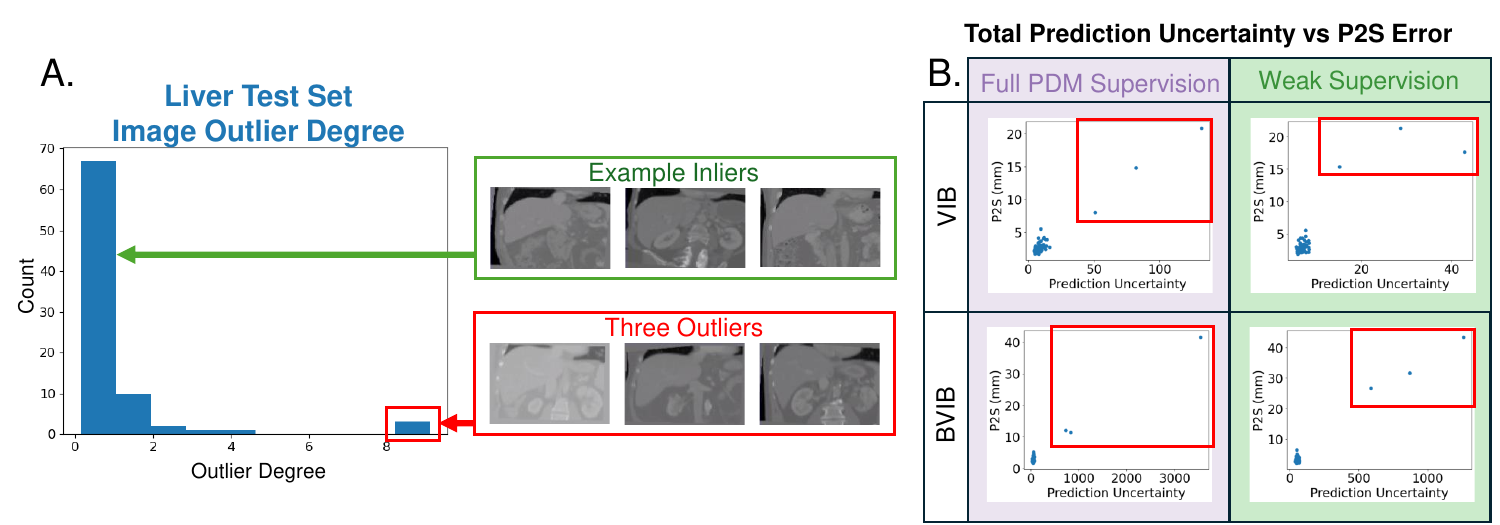}
    \end{center}
    \caption{Liver outlier detection results. The histogram plots the distribution of image outlier degrees across the liver test set, highlighting the three outliers. Slices of inlier and outlier images are shown. The total prediction uncertainty and P2S error (averaged across each shape) are plotted across for each model. The three outlier cases have high uncertainty and are highlighted in red boxes.}
    \label{fig:liver_outliers}
\end{figure*}

Fig. \ref{fig:liver_outliers} shows the distribution of image outlier degrees across the randomly selected liver test set. Three outlier cases are identified in this histogram. These three cases are also clearly identifiable in the P2S error vs prediction uncertainty scatter plots in Fig. \ref{fig:liver_outliers}. Here, prediction uncertainty is the total aleatoric for VIB models and the sum of the total aleatoric and epistemic estimates for BVIB. The outlier cases are clearly identifiable, given the prediction uncertainty resulting from both full and weak supervision, suggesting the proposed weakly supervised approach is not detrimental to OOD detection.

\section{Conclusion}

We proposed an alternative training approach to BVIB-DeepSSM with reduced supervision. The proposed framework matches PDM-supervision accuracy while significantly streamlining the training pipeline. 
In future work, the point cloud shape representations could also be leveraged to learn a more expressive prior $p(\z)$. 
In \cite{xu2019necessity}, it is proven that learning the variational autoencoder (VAE) latent prior is necessary for reaching the extremum of the VAE objective. This proof can be directly applied to show learning $p(\z)$ is necessary for reaching the extremum of the VIB objective.
Future work could explore utilizing a point cloud autoencoder to learn $p(\z)$ in a shape-informed manner. 
Overall, this work improves the feasibility of SSM construction from images, making SSM more accessible as a tool for clinical research. 

\section{Acknowledgements}
This work was supported by the National Institutes of Health under grant numbers NIBIB-U24EB029011, NIAMS-R01AR076120, NHLBI-R01HL135568, and NIBIB-R01EB016701. We thank the University of Utah Division of Cardiovascular Medicine for providing left atrium MRI scans and segmentations from the Atrial Fibrillation projects and the ShapeWorks team.

%
%
\bibliographystyle{splncs04}
\bibliography{refs}
\end{document}